\documentclass[wcp]{jmlr}

\usepackage{longtable}

\usepackage{booktabs}

\usepackage{lineno}

\makeatletter
\let\Ginclude@graphics\@org@Ginclude@graphics 
\makeatother

\jmlrvolume{}
\jmlryear{2023}
\jmlrworkshop{ACML 2023}

\usepackage{mathtools}

\usepackage{hyperref}

\usepackage{caption}
\usepackage{graphicx}
\usepackage[symbol]{footmisc}
\usepackage{cancel}
\hypersetup{colorlinks,citecolor=blue,linkcolor=red,urlcolor=blue}
\usepackage{mathtools}
\usepackage{booktabs} 
\usepackage{soul}
\sethlcolor{red}
\newcommand{\mathcolorbox}[2]{\colorbox{#1}{$\displaystyle #2$}}

\colorlet{myred}{red!40!white}
\colorlet{myblue}{blue!40!white}
\colorlet{mygreen}{green!40!white}
\usepackage{mathrsfs}

\newcommand{\bE}{\boldsymbol{\mathrm{E}}}
\newcommand{\bX}{\boldsymbol{\mathrm{X}}}

\newcommand{\bx}{\boldsymbol{x}}
\newcommand{\by}{\boldsymbol{y}}

\newcommand{\bw}{\boldsymbol{\mathrm{w}}}

\newcommand{\bH}{\boldsymbol{\mathrm{H}}}

\newcommand{\bI}{\boldsymbol{\mathrm{I}}}

\newcommand{\bmu}{\boldsymbol{\mu}}

\newcommand{\bLambda}{\boldsymbol{\Lambda}}

\usepackage{tcolorbox}
\newtcolorbox{mybox}[3][]
{
  colframe = #2!25,
  colback  = #2!10,
  coltitle = #2!20!black,
  #1,
}

\DeclareMathOperator*{\argmax}{arg\,max}
\DeclareRobustCommand{\violet}{%
  \begingroup\normalfont
  \includegraphics[height=\fontcharht\font`\B]{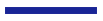}%
  \endgroup
}
\DeclareRobustCommand{\lightviolet}{%
  \begingroup\normalfont
  \includegraphics[height=\fontcharht\font`\B]{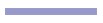}%
  \endgroup
}
\DeclareRobustCommand{\pink}{%
  \begingroup\normalfont
  \includegraphics[height=\fontcharht\font`\B]{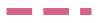}%
  \endgroup
}
\DeclareRobustCommand{\blue}{%
  \begingroup\normalfont
  \includegraphics[height=\fontcharht\font`\B]{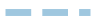}%
  \endgroup
}

\usepackage[capitalize,noabbrev]{cleveref}

\usepackage[textsize=tiny]{todonotes}

\title[The Fine Print on Tempered Posteriors]{The Fine Print on Tempered Posteriors}

\author{\Name{Konstantinos Pitas} \Email{pitas.konstantinos@inria.fr} \\ \and
\Name{Julyan Arbel} \Email{julyan.arbel@inria.fr}\\
\addr Univ. Grenoble Alpes, Inria, CNRS, Grenoble INP, LJK, 38000 Grenoble, France}

\editors{Berrin Yan{\i}ko\u{g}lu and Wray Buntine}

\begin{document}

\maketitle

\begin{abstract}
We conduct a detailed investigation of tempered posteriors and uncover a number of crucial and previously undiscussed points. Contrary to previous results, we first show that for realistic models and datasets and the tightly controlled case of the Laplace approximation to the posterior, stochasticity does not in general improve test accuracy. The coldest temperature is often optimal. One might think that Bayesian models with some stochasticity can at least obtain improvements in terms of calibration. However, we show empirically that when gains are obtained this comes at the cost of degradation in test accuracy. We then discuss how targeting Frequentist metrics using Bayesian models provides a simple explanation of the need for a temperature parameter $\lambda$ in the optimization objective. Contrary to prior works, we finally show through a PAC-Bayesian analysis that the temperature $\lambda$ cannot be seen as simply fixing a misspecified prior or likelihood.
\end{abstract}
\begin{keywords}
Tempered Posteriors; Cold Posteriors; PAC-Bayes
\end{keywords}

\section{Introduction}
An influential line of work starting from \citet{wenzel2020good} highlighted that Bayesian neural networks typically exhibit better test-time predictive performance if the posterior distribution is ``sharpened'' through tempering. Tempering with a temperature $\lambda$ is either applied to both the data likelihood and the prior resulting in cold posteriors or only to the data likelihood resulting in tempered posteriors. The empirical need for this tempering highlights a fundamental mismatch between Bayesian theory and practice. As the number of training samples increases, Bayesian theory states that the posterior distribution should be concentrating more and more on the true model parameters, under mild regularity assumptions. At any time, the posterior is our best guess at the true model parameters, without having to resort to heuristics.

The experimental setups where the need for tempering arises have been hard to pinpoint precisely. 
Likelihood misspecification has proven to be a prominent explanation. \citet{nabarro2021data, bachmann2022tempering,kapoor2022uncertainty,aitchison2020statistical} argue that the aleatoric uncertainty in the likelihood function becomes misspecified in many experiments due to data augmentation and/or due to curation of the data. A common thread in the above works is that a correct likelihood would lead to optimal test performance and tempering would be unnecessary. In the original paper \citet{wenzel2020good} as well as subsequent ones \citep{nabarro2021data, bachmann2022tempering,kapoor2022uncertainty,aitchison2020statistical} another conclusion is that Bayesian neural networks outperform deterministic ones in test metrics for the correct value of $\lambda$. 

We challenge both theoretically and experimentally the above assumptions and conclusions. In a strict sense, we will be focusing on tempered posteriors, while noting that for Gaussian priors and posteriors the two can be seen as equivalent \citep{aitchison2020statistical}.  We make the following four contributions.

We first observe that in previous works \citep{nabarro2021data, bachmann2022tempering,kapoor2022uncertainty,aitchison2020statistical} even after ``fixing" the likelihood with various approaches, the test accuracy seems to be improving as the temperature is decreasing, though it is implied that some stochasticity is beneficial. In the tightly controlled case of the Laplace approximation, we show experimentally that in terms of test accuracy, stochasticity does not help (Section \ref{experiments:fine}). In contrast to previous works, our conclusion is that the \emph{deterministic} network is nearly always optimal. 

Second, one can question if such an evaluation is fair for Bayesian models. Indeed practitioners often motivate the use of Bayesian models so as to improve the calibration of predictions. We  empirically show that for the negative log-likelihood (NLL) and the expected calibration error (ECE), multiple values of $\lambda$ are optimal, i.e. some stochasticity helps (Section \ref{experiments:fine}). Crucially though, improving the test ECE comes in general at the cost of reducing test accuracy.
%

Third, given the poor performance of purely Bayesian models for Frequentist metrics, we then posit that there exists a simpler explanation for cold and tempered posteriors, specifically that Bayesian inference \textit{does not readily provide high-probability guarantees on out-of-sample performance} (Section \ref{simple_explanation}). Existing theorems simply describe \emph{posterior contraction} to the true posterior \citep{ghosal2000convergence,blackwell1962merging}. However,  practitioners \citep{nabarro2021data, bachmann2022tempering,kapoor2022uncertainty,aitchison2020statistical} evaluate posteriors on Frequentist test metrics such as test accuracy. This discrepancy between using Bayesian models and evaluating on Frequentist metrics is enough to justify the use of a temperature parameter. To prove out-of-sample performance with high probability, one needs a generalization bound such as a PAC-Bayes bound \citep{mcallester1999some,catoni2007pac,alquier2016properties,dziugaite2017computing}. The tightest of these bounds \citep{catoni2007pac} naturally incorporates a temperature parameter $\lambda$. 

Finally, we investigate the question of whether tempered posteriors can be seen as fixing a misspecified likelihood (Section \ref{sec:effect}). With a simplified PAC-Bayesian analysis, we argue that tempered posteriors do not only compensate for a misspecified aleatoric uncertainty, contrary to \citet{nabarro2021data, bachmann2022tempering,kapoor2022uncertainty,aitchison2020statistical}. Our bound implies that the specific weight instantiation of a neural network influences the choice of $\lambda$. We confirm this experimentally for the case of the Laplace approximation. We use different MAP estimates to fit the Laplace approximation. For a fixed value of $\lambda$, the test accuracy changes even though \emph{both the likelihood function and the inherent aleatoric uncertainty of the data remain the same.}

We also include a detailed FAQ section in the Appendix, together with all the proofs and additional experiments.
\section{Related work}\label{related_work}
\citet{aitchison2020statistical} argues that tempering results from the fact that common computer vision benchmarks are curated. That is, each label has been independently verified by multiple labelers, thus the aleatoric uncertainty for all labels is low. He proposes a generative model that implies a tempered likelihood that accounts for data curation. \citet{kapoor2022uncertainty} describe the softmax-cross-entropy loss as resulting from a multinomial likelihood function together with a Dirichlet prior distribution. They propose a new likelihood and a corresponding Dirichlet prior which explicitly models aleatoric uncertainty through an additional hyper-parameter. They argue that a softmax-cross-entropy loss does not explicitly allow tuning the aleatoric uncertainty of the data, thus necessitating tempered posteriors.  \citet{bachmann2022tempering} argue that data augmentation reduces the effective number of training samples. Thus tempering reduces the nominal number of training samples so as to match the effective number. \citet{nabarro2021data} explore whether data-augmentation \emph{increases} the effective sample size. They propose a new valid likelihood that takes into account data augmentations and discover that the cold-posterior effect persists. Despite the above explanations, the need for tempering persists even in settings without data curation and augmentation, and when the likelihood is suitably modified.

Certifying generalization to out-of-sample data has been extensively explored for the case of deep neural networks \citep{dziugaite2021role,dziugaite2017computing,zhou2018non,lotfi2022pac,bartlett2017spectrally}. The current state-of-the-art bounds \citep{dziugaite2021role,lotfi2022pac} utilize the Catoni PAC-Bayes bound \cite{catoni2007pac}, together with a data-dependent prior, and find non-vacuous and in some cases tight bounds. We base part of our arguments on the Catoni bound.

The posterior contraction properties of Bayesian inference have been extensively studied \citep{ghosal2000convergence,blackwell1962merging}. Of note are results that show contraction even in the case of mispecification \citet{grunwald2012safe,grunwald2017inconsistency,grunwald2007suboptimal,bhattacharya2019bayesian}, using tempered posteriors restricted to $\lambda<1$ (warm posteriors). 

\citet{grunwald2007suboptimal,grunwald2012safe,grunwald2017inconsistency} explore Safe-Bayes, which finds a temperature parameter $\lambda$ by taking a sequential view of Bayesian inference. They find a Cèsaro averaged posterior, which is an average of the posteriors at different optimization steps, and which does not coincide with the standard posterior. The Safe-Bayes analysis is also restricted to the case where $\lambda<1$. By contrast we provide an analytical expression of the bound on true risk, given $\lambda$, and also numerically investigate the case of $\lambda>1$. Our analysis thus provides intuition regarding which parameters (for example the curvature) might result in tempering. \citet{catoni2007pac} discusses the optimal value of the temperature $\lambda$ for PAC-Bayes bounds, for \emph{fixed} priors and posteriors. By contrast we investigate the case where the posterior is optimized for different $\lambda$ and which is the relevant one for the tempering literature. 
\citet{bhattacharya2019bayesian} derive a general form of the PAC-Bayes inequality albeit to analyze posterior contraction. In several prediction settings, out-of-sample performance can however be established directly using this result. 

Modern contraction analyses and PAC-Bayes bounds are closely linked \citep{bhattacharya2019bayesian,grunwald2012safe}. It is thus important to base our results on the tightest analyses available for the deep learning setting. As previously mentioned, for Bayesian neural networks the particular Catoni form of the PAC-Bayesian inequality has been empirically shown to result in state-of-the-art bounds. We, therefore, use the Catoni bound in our analyses where applicable. 

\section{Limitations}
We base our analysis on the Laplace approximation to the posterior. Therefore the question arises as to whether our results generalize to other inference methods. Here we simply note that our empirical results align with existing results in the MCMC case such as \cite{kapoor2022uncertainty, nabarro2021data, noci2021disentangling}. A further limitation is that PAC-Bayes bounds are sometimes (relatively) tight but more often they are loose. This brings into question the practical utility of a PAC-Bayes theoretical analysis.

\section{Stochasticity typically hurts performance}\label{experiments:fine}
We first explore whether any stochasticity improves the test accuracy, compared to a deterministic network. Here, working in the Laplace approximation setting is crucial. Previous studies dealt with MCMC methods and have not adequately explored what happens when the temperature becomes ``frozen". The case of \citet{wenzel2020good} is striking as the ``frozen" network and the deterministic network test accuracies do not match \citep[Figure 1 of ][]{wenzel2020good}. The Laplace approximation has a simple intuitive behaviour: the posterior collapses on a Dirac delta at the MAP as $\lambda\rightarrow+\infty$. We then investigate how $\lambda$ affects the other most popular test metrics (NLL and ECE). In particular we explore whether one can improve the ECE without hurting test accuracy, which is the most pertinent question for practitioners.
\subsection{Definitions}
We denote the learning sample $(X,Y)=\{(\bx_i,y_i)\}^n_{i=1}\in(\mathcal{X}\times\mathcal{Y})^n$, that contains $n$ input-output pairs. Observations $(X,Y)$ are assumed to be sampled randomly from a distribution $\mathcal{D}$. Thus, we denote $(X,Y)\sim\mathcal{D}^n$ the i.i.d observation of $n$ elements. We consider loss functions $\ell:\mathcal{F}\times\mathcal{X}\times\mathcal{Y}\rightarrow\mathbb{R}$, where $\mathcal{F}$ is a set of predictors $f:\mathcal{X}\rightarrow\mathcal{Y}$. We also denote the risk  $\mathcal{L}^{\ell}_{\mathcal{D}}(f)=\bE_{(\bx,y)\sim\mathcal{D}}\ell(f,\bx,y)$ and the empirical risk $\hat{\mathcal{L}}^{\ell}_{X,Y}(f)=(1/n)\sum_i\ell(f,\bx_i,y_i)$. We consider two probability measures, the prior $\pi\in\mathcal{M}(\mathcal{F})$ and the approximate posterior $\hat{\rho}\in\mathcal{M}(\mathcal{F})$. Here, $\mathcal{M}(\mathcal{F})$ denotes the set of all probability measures on $\mathcal{F}$.  We encounter cases where we make predictions using the approximate posterior predictive distribution $\bE_{f\sim\hat{\rho}}[p(y|\bx,f)]$. 
We will use two loss functions, the non-differentiable zero-one loss $\ell_{01}(f,\bx,y)=\mathbb{I}(\argmax_{j}f(\bx)_j\neq y)$, and the negative log-likelihood (NLL), which is a commonly used differentiable surrogate $\ell_{\text{nll}}(f,\bx,y)=-\log(p(y|\bx,f))$. We assume that outputs of $f$ form a probability distribution $p(y|\bx,f)$ either through a Gaussian likelihood (in the case of regression) or using the softmax activation function (in the case of classification). Given the above, the Evidence Lower Bound (ELBO) has the following form
\begin{equation}\label{generalized_obj}
-\bE_{f\sim\hat{\rho}}\hat{\mathcal{L}}^{\ell_{\mathrm{nll}}}_{X,Y}(f)-\frac{1}{\lambda n}\mathrm{KL}(\hat{\rho}\Vert \pi),
\end{equation}
where $\lambda=1$. Note that our temperature parameter $\lambda$ is the \emph{inverse} of the parameter $T=1/\lambda$ typically used in tempered posterior works. Here, cold posteriors are the result of $\lambda>1$. 
Our setup is discussed in \citet{wenzel2020good}, p3 Section 2.3, and used in \citet{bachmann2022tempering,aitchison2020statistical,kapoor2022uncertainty}. While \citet{wenzel2020good} use MCMC to conduct their experiments, we opt for the ELBO for analytical tractability. 

\subsection{Experimental setup: Laplace approximation}
The ELBO \eqref{generalized_obj} is minimized at the probability density $\rho^{\star}(f)$ given by \citet{catoni2007pac}:
$$
\rho^{\star}(f) \coloneqq \pi(f)e^{-\lambda n \hat{\mathcal{L}}^{\ell_{\mathrm{nll}}}_{X,Y}(f)}/\bE_{f\sim\pi}\left[ e^{-\lambda n \hat{\mathcal{L}}^{\ell_{\mathrm{nll}}}_{X,Y}(f)}\right]. 
$$
We will use the Laplace approximation to the posterior in our experiments. This is equivalent to approximating $\lambda n\hat{\mathcal{L}}^{\ell_{\mathrm{nll}}}_{X,Y}(f)$ using a second order Taylor expansion around a minimum $\bw_{\hat{\rho}}$, such that 
\begin{equation}
\lambda n\hat{\mathcal{L}}^{\ell_{\mathrm{nll}}}_{X,Y}(f_{\bw})\approx\lambda n\hat{\mathcal{L}}^{\ell_{\mathrm{nll}}}_{X,Y}(f_{\bw_{\hat{\rho}}})
+\frac{\lambda}{2} (\bw-\bw_{\hat{\rho}})^{\top}
\bH 
(\bw-\bw_{\hat{\rho}}),
\end{equation}
where $\bH$ is the network Hessian  $n\nabla\nabla\hat{\mathcal{L}}^{\ell_{\mathrm{nll}}}_{X,Y}(f_{\bw})\vert_{\bw=\bw_{\hat{\rho}}}$. 
Assuming a Gaussian prior $\pi = \mathcal{N}(0,\sigma_{\pi}^2\bI)$, the Laplace approximation to the posterior $\hat{\rho}$ is again Gaussian 
$$\hat{\rho}=\mathcal{N}\big(\bw_{\hat{\rho}},\left(\lambda\bH+ \bI/\sigma_{\pi}^2\right)^{-1}\big)$$.

The Hessian is generally infeasible to compute in practice for modern deep neural networks, such that many approaches employ the generalized Gauss--Newton (GGN) approximation $\bH^{\mathrm{GGN}}\coloneqq \sum_{i=1}^n \mathcal{J}_{\bw}(\bx_i)^{\top}\bLambda(\by_i;f_i)\mathcal{J}_{\bw}(\bx_i),$
where $\mathcal{J}_{\bw}(\bx)$ is the network per-sample Jacobian $\left[\mathcal{J}_{\bw}(\bx)\right]_{c}=\nabla_{\bw}f_c(\bx;\bw_{\hat{\rho}})$, and $\bLambda(\by;f)=-\nabla^2_{ff}\log p(\by;f)$ is the per-input noise matrix \citep{kunstner2019limitations}. We will use two simplified versions of the GGN: (i) an isotropic approximation with variance $\sigma^2_{\hat{\rho}}(\lambda)$ equal to $\left(\frac{\lambda h}{d}+\frac{1}{\sigma_{\pi}^2}\right)^{-1}$, and $h$ equal to 
    $\mathrm{tr}(\bH^{\mathrm{GGN}})= \sum_{i,j,k}[\bLambda(\by_i;f)]_{kk}(\nabla_{\bw}f_k(\bx_i;\bw_{\hat{\rho}})_j)^2$, and
    (ii) the Kronecker-Factorized Approximate Curvature (KFAC) approximation \citep{martens2015optimizing} which only retains a block-diagonal part of the GGN.

\subsection{Experimental setup: training}
\begin{figure*}[t!]
\centering
  \includegraphics[width=\textwidth]{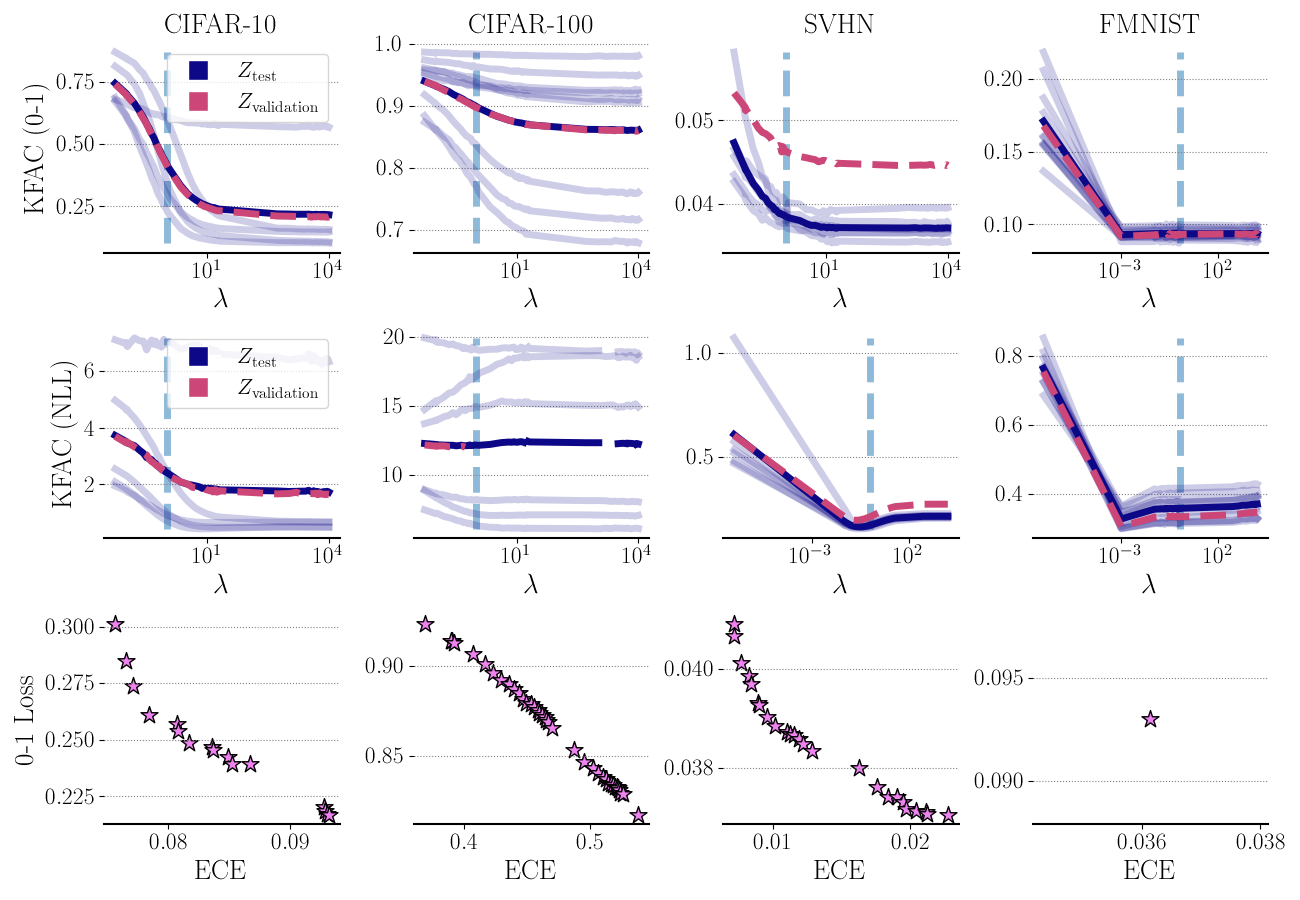}
  \caption{Test 0-1 loss mean \violet, as well as 10 MAP trials \lightviolet,  along with the mean of the validation set \pink (we denote $\lambda=1$ by \blue): KFAC Laplace 0-1 loss (top) and KFAC NLL (middle) and 0-1 loss vs ECE (bottom row). For the standard KFAC posteriors the test and validation 0-1 loss have a rapid improvement as $\lambda\uparrow$ followed by a plateau. Coldest posteriors $\lambda\gg1$ are almost always best. The NLL has more complex behaviour, with cold and warm posteriors depending on the dataset. There is a clear tradeoff between 0-1 loss and ECE.}
  \label{exp_fig:KFAC_main_paper}
\end{figure*}
When making predictions, we use the posterior predictive distribution $\bE_{\bw\sim\hat{\rho}}[p(y|\bx,f_{\bw})]$ of the \emph{full neural network model}, meaning that samples from $\hat{\rho}$ are inputted to the full neural network (as opposed to, for example, its linearization). Since the 0-1 loss is not differentiable, the posterior estimated with the cross-entropy loss will be used for classification problems.

We have tested extensively in realistic classification tasks. We used the CIFAR-10, CIFAR-100 \citep{krizhevsky2009learning}, SVHN \citep{netzer2011reading} and FashionMnist \citep{xiao2017/online} datasets. 
In all experiments, we split the dataset into three sets. These three are the typical prediction tasks sets: training set $Z_{\mathrm{train}}$, testing set $Z_{\mathrm{test}}$, and validation set $Z_{\mathrm{validation}}$. 
We use Monte Carlo sampling with 100 samples to make predictions. For the CIFAR-10, CIFAR-100, and SVHN datasets, we use a WideResNet22 \citep{zagoruyko2016wide}, with Fixup initialization \citep{zhang2019fixup}. For the FashionMnist dataset, we use a convolutional architecture with three convolutional layers, followed by two fully connected non-linear layers. More details on the experimental setup can be found in the Appendix.

We find ten MAP estimates for the neural network weights of the CIFAR-10, CIFAR-100, SVHN and FMNIST datasets by training on $Z_{\mathrm{train}}$ using SGD. 
We then fit a KFAC Laplace approximation to each MAP estimate using $(X,Y)=Z_{\mathrm{train}}$ and also choose the prior through optimizing the marginal likelihood. We estimate the 0-1 loss, negative log-likelihood (NLL), and Expected Calibration Error (ECE) of the posterior predictive on $Z_{\mathrm{test}}$ and $Z_{\mathrm{validation}}$.

\subsection{Observations}
We first test a setting without data augmentation. We plot the results for the 0-1 loss in Figure \ref{exp_fig:KFAC_main_paper} (top row). \emph{Contrary to \citet{wenzel2020good} in terms of test 0-1 loss, the MAP estimate obtained where $\lambda\gg 1$ and the posterior is ``coldest" is almost always optimal}. This result is more coherent than the one in \citet{wenzel2020good} (Figure 1) where the coldest temperature 0-1 test loss and the MAP 0-1 test loss don't match. It highlights that in a tightly controlled setting, Bayesian approaches often don't improve at all over deterministic ones in terms of test Frequentist metrics such as the 0-1 loss, \emph{for any temperature $\lambda$}.

We then plot in Figure~\ref{exp_fig:KFAC_main_paper} (middle row) the results for the NLL. Even without data augmentation and even when we optimize the prior variance using the marginal likelihood, we find that all three cases of temperatures (cold posterior, warm posterior, as well as posterior with $\lambda =1$) can be optimal, for varying datasets. \emph{This highlights the importance of the choice of the evaluation metric when discussing the cold posterior effect, as results can vary significantly depending on our choice.} 

We then elaborate on this finding and further ask the crucial question: \textit{If the Laplace approximation does not improve the test 0-1 loss in general, can it retain (for some value of $\lambda$) the same test 0-1 loss while improving calibration?} In Figure \ref{exp_fig:KFAC_main_paper}, we see that in most of our experiments, we couldn't find such a temperature $\lambda$. Apart from FMNIST, \emph{there seems to be a clear tradeoff between test 0-1 loss and calibration error.}

We repeat the experiment for the case of data augmentation, for the CIFAR-10 and CIFAR-100 datasets. We use the standard augmentations of random crops and rotations. We perform augmentation both when estimating the MAP, as well as when fitting the Laplace approximation. 
We include the results in the Appendix. We see similar behavior to the setting without data augmentation. Our only additional observation is that test accuracy has improved for both CIFAR-10 and CIFAR-100 (without augmentation not only does training suffer but numerical issues arise when performing matrix inversion for the Laplace approximation). 

\begin{mybox}{blue}{Summary}
Empirically (for the Laplace case) the coldest temperature is almost always optimal in terms of test accuracy. Even though one can improve calibration for a given temperature, this comes at the cost of accuracy.
\end{mybox}
\section{Tempered Posteriors: A simple explanation}\label{simple_explanation}
In light of the generally poor performance of purely Bayesian methods in terms of test Frequentist metrics for deep learning, we explore a simple explanation regarding the need for tempering: the mismatch between Bayesian guarantees and practice. We start by noting that in supervised prediction, what we often try to minimize is
\begin{equation}
    \label{eq:eval-metric-classif}
    \mathrm{KL}(p_{\mathcal{D}}(y|\bx)\Vert \bE_{f\sim\hat{\rho}}[p(y|\bx,f)])
    =\bE_{\bx,y\sim\mathcal{D}}\left[\ln\frac{p_{\mathcal{D}}(y|\bx)}{\bE_{f\sim\hat{\rho}}[p(y|\bx,f)]}\right],
\end{equation} 
the conditional relative entropy \citep{cover1999elements} between the true conditional distribution $p_{\mathcal{D}}(y|\bx)$ and the posterior predictive distribution $\bE_{f\sim\hat{\rho}}[p(y|\bx,f)]$. For example, this is implicitly the  minimized quantity when optimizing classifiers with  the cross-entropy loss \citep{masegosa2019learning,morningstar2022pacm}. It is also on this and similar predictive metrics that the tempering effect appears. In the following, we will outline the relationship between the ELBO, PAC-Bayes, and \eqref{eq:eval-metric-classif}. 
\subsection{ELBO}
We assume a training sample $(X,Y)\sim\mathcal{D}^n$ as before, denote $p(\bw|X,Y)$ the true posterior probability over predictors $f$ parameterized by $\bw$ (typically weights for neural networks), and $\pi$ and $\hat{\rho}$ respectively the prior and variational posterior distributions as before. The ELBO has the following form
\begin{equation*}
\mathrm{KL}(\hat{\rho}(\bw)\Vert p(\bw|X,Y))
=-n\textcolor{orange}{\underbrace{\left(-\bE_{f\sim\hat{\rho}}\hat{\mathcal{L}}_{X,Y}^{\ell_{\mathrm{nll}}}(f)-\frac{1}{n}\mathrm{KL}(\hat{\rho}\Vert \pi)\right)}_{\mathrm{ELBO}}}+\ln p(Y|X).
\end{equation*}
Thus, maximizing the ELBO can be seen as minimizing the KL divergence between the true posterior and the variational posterior over the weights $\mathrm{KL}(\hat{\rho}(\bw)\Vert p(\bw|X,Y))$. The true posterior distribution $p(\bw|X,Y)$ gives more probability mass to predictors which are more likely given the training data, \emph{however these predictors do not provably minimize }$\mathrm{KL}(p_{\mathcal{D}}(y|\bx)\Vert \bE_{f\sim\hat{\rho}}[p(y|\bx,f)])$, the evaluation metric of choice for supervised prediction. 
Similar issues exist with theorems on posterior contraction. These typically guarantee that the posterior concentrates on the most probable set of parameters, in the well-specified regime 
\citep{blackwell1962merging}. 
When we are not interested in inferring the most probable parameters but in predicting on new data, it is important to derive a certificate of generalization through a generalization bound, which directly bounds out-of-sample performance. 
In the following, we focus on analyzing a PAC-Bayes bound in order to obtain insights into when tempering is necessary. As discussed in Section \ref{related_work}, modern contraction analyses are closely related to PAC-Bayes, and one can sometimes move from a contraction result to a generalization bound. However, the specific forms discussed in the following section currently provide state-of-the-art generalization bounds for modern deep neural networks \citep{dziugaite2021role,lotfi2022pac}.
\subsection{PAC-Bayes} 
We first look at the following bound denoted by $\mathcal{B}_{\mathrm{Alquier}}$. It was considered by \citet{alquier2016properties} (Theorem 4.1); see also Theorem 1 by \citet{masegosa2019learning} for a statement under the same conditions. We will see that it is natural to assume a temperature $\lambda$ when trying to bound out-of-sample performance, as opposed to guaranteeing posterior contraction.
\begin{theorem}[$\mathcal{B}_{\mathrm{Alquier}}$, \citealp{alquier2016properties}]\label{th:alquier}
Given a distribution $\mathcal{D}$ over $\mathcal{X}\times\mathcal{Y}$, a hypothesis set $\mathcal{F}$, a loss function $\ell:\mathcal{F}\times\mathcal{X}\times\mathcal{Y}\rightarrow\mathbb{R}$, a prior distribution $\pi$ over $\mathcal{F}$, real numbers $\delta \in (0,1]$ and $\lambda>0$, with probability at least $1-\delta$ over the choice $(X,Y)\sim\mathcal{D}^n$, we have for all $\hat{\rho}$ on $\mathcal{F}$
\begin{equation*}
\bE_{f\sim\hat{\rho}}\mathcal{L}_{\mathcal{D}}^{\ell}(f)\leq \bE_{f\sim\hat{\rho}}\hat{\mathcal{L}}_{X,Y}^{\ell}(f)\\
+\frac{1}{\lambda n}\left[\mathrm{KL}(\hat{\rho}\Vert \pi)+\ln\frac{1}{\delta}   + \Psi_{\ell,\pi,\mathcal{D}}(\lambda,n) \right]
\end{equation*}
where $\Psi_{\ell,\pi,\mathcal{D}}(\lambda,n)$ is equal to 
$\ln \bE_{f\sim\pi}\bE_{X',Y'\sim\mathcal{D}^n}\exp \left[\lambda n\left(\mathcal{L}_{\mathcal{D}}^{\ell}(f)- \hat{\mathcal{L}}_{X',Y'}^{\ell}(f)\right)  \right]$.
\end{theorem}
\begin{figure*}[t!]
\centering
  \includegraphics[width=\textwidth]{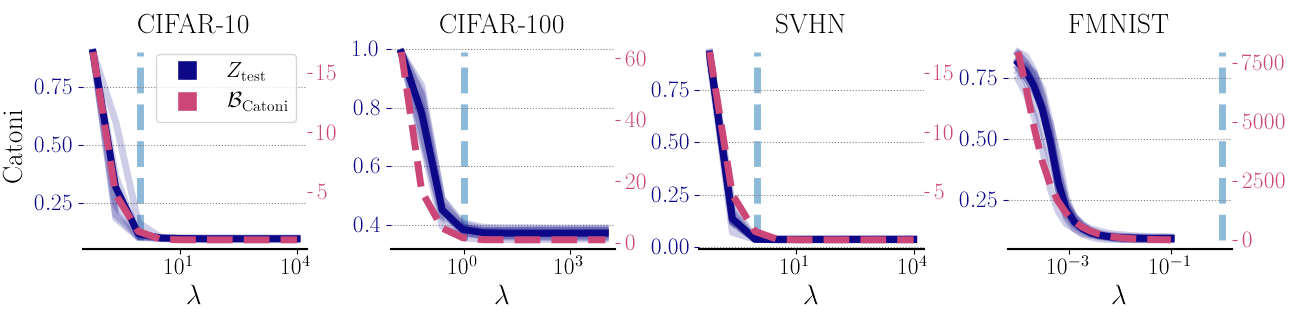}
  \caption{We plot the Catoni bound, as well as the test misclassification error for the CIFAR-10, CIFAR-100, SVHN and FMNIST datasets and the Isotropic Laplace approximation. We see that the Catoni bounds tightly correlate with the test error. As the temperature $\lambda$ increases, the bound predicts a smaller test error, this is reflected in the empirical data as the need for a tempered posterior.
  }
  \label{catoni_bound_main_fig}
\end{figure*}
%
There are three different terms in the above bound. The empirical risk term $\bE_{f\sim\hat{\rho}}\hat{\mathcal{L}}_{X,Y}^{\ell}(f)$ is the empirical mean of the loss of the classifier over all training samples. The KL term $1/(\lambda n)\mathrm{KL}(\hat{\rho}\Vert \pi)$ is the complexity of the model, which in this case is measured as the KL-divergence between the posterior and prior distributions. The Moment term $1/(\lambda n)\Psi_{\ell,\pi,\mathcal{D}}(\lambda,n)$ is the log-Laplace transform of the difference between the risk and empirical risk for a reversal of the temperature. We can typically make assumptions on the tails of the loss to ensure that the Moment term is bounded \citep[see eg][]{germain2016pac}. 
Using Theorem \ref{th:alquier} together with Jensen's inequality, one can bound \eqref{eq:eval-metric-classif} directly as follows
\begin{equation*}
\begin{split}
&\mathrm{KL}(p_{\mathcal{D}}(y|\bx)\Vert \bE_{f\sim\hat{\rho}}[p(y|\bx,f)])\\
&=\bE_{\bx,y\sim\mathcal{D}}[-\ln\bE_{f\sim\hat{\rho}}[p(y|\bx,f)]]+\bE_{\bx,y\sim\mathcal{D}}[\ln p_{\mathcal{D}}(y|\bx)]\\
&\leq\bE_{\bx,y\sim\mathcal{D}}[\bE_{f\sim\hat{\rho}}[-\ln p(y|\bx,f)]]+\bE_{\bx,y\sim\mathcal{D}}[\ln p_{\mathcal{D}}(y|\bx)]\\
&\leq\textcolor{orange}{\underbrace{\bE_{f\sim\hat{\rho}}\hat{\mathcal{L}}_{X,Y}^{\ell_{\mathrm{nll}}}(f)+\frac{1}{\lambda n}\left[\mathrm{KL}(\hat{\rho}\Vert \pi)+\ln\frac{1}{\delta}+ \Psi_{\ell_{\mathrm{nll}},\pi,\mathcal{D}}\right]}_{\mathrm{PAC}\text{-}\mathrm{Bayes} }}+\bE_{\bx,y\sim\mathcal{D}}[\ln p_{\mathcal{D}}(y|\bx)].
\end{split}
\end{equation*}
%
The last line holds under the conditions of Theorem~\ref{th:alquier} and in particular with probability at least $1-\delta$ over the choice $(X,Y)\sim\mathcal{D}^n$. Notice here the presence of the temperature parameter $\lambda\geq0$, which need not be $\lambda=1$. 
%

    \emph{It is easy to see that maximizing the ELBO is equivalent to minimizing a PAC-Bayes bound for $\lambda=1$, which might not necessarily be optimal for a finite sample size. More specifically even for exact inference
    the Bayesian posterior predictive distribution does not necessarily minimize $\mathrm{KL}(p_{\mathcal{D}}(y|\bx)\Vert \bE_{f\sim\hat{\rho}}[p(y|\bx,f)])$.}

%
We note that the temperature is inherent in the derivation of PAC-Bayes bounds \citep{begin2016pac} (see also Appendix). It more generally occurs when trying to apply Chernoff bounds on the tails of random variables. It is necessary here because we are trying to bound out-of-sample performance, with high probability.

\subsection{Classification tasks}
For classification tasks, we are typically mainly interested in achieving low expected zero-one risk 
$
\bE_{f\sim\hat{\rho}}\mathcal{L}^{\ell_{\mathrm{01}}}_{\mathcal{D}}(f).
$
The ELBO objective is not directly related to this risk. However in the PAC-Bayesian literature, there exist bounds specifically adapted to it. In the following, we will use one of the tightest and most commonly used bounds, the ``Catoni" bound, denoted $\mathcal{B}_{\mathrm{Catoni}}$ from \citet{catoni2007pac} Theorem 1.2.6.   
\begin{theorem}[$\mathcal{B}_{\mathrm{Catoni}}$, \citealp{catoni2007pac}]\label{th:catoni}
Given a distribution $\mathcal{D}$ over $\mathcal{X}\times\mathcal{Y}$, a hypothesis set $\mathcal{F}$, the 0-1 loss function $\ell_{01}:\mathcal{F}\times\mathcal{X}\times\mathcal{Y}\rightarrow[0,1]$, a prior distribution $\pi$ over $\mathcal{F}$, a real number $\delta\in(0,1]$, and a real number $\lambda>0$,  with $\Phi^{-1}_\lambda(x) = \frac{1-e^{-\lambda x}}{1-e^{-\lambda}}$, we have  with probability at least $1-\delta$ over $(X,Y)\sim\mathcal{D}^n$
\begin{equation*}
\forall{\hat{\rho}} \;\mathrm{on}\; \mathcal{F}:\bE_{f\sim\hat{\rho}} \mathcal{L}^{\ell_{01}}_{\mathcal{D}}(f) \leq\\ 
\Phi^{-1}_\lambda\left(\bE_{f\sim\hat{\rho}}\hat{\mathcal{L}}^{\ell_{01}}_{X,Y}(f)+\frac{1}{\lambda n}\left[\mathrm{KL}(\hat{\rho}||\pi)+\ln{\frac{1}{\delta}}\right]\right).
\end{equation*}
\end{theorem}
Similarly to the Alquier bound, the empirical risk term is the empirical mean of the loss of the classifier over all training samples. The KL term is the complexity of the model, which in this case is measured as the KL divergence between the posterior and prior distributions. The Moment term has been absorbed in this case in the $\Phi^{-1}_\lambda$ function.

\subsection{Experiments}

PAC-Bayes bounds require correct control of the prior mean as the $\ell_2$ distance between prior and posterior means in the KL term is often the dominant term in the bound. To control this distance, we follow a variation of the approach in \citet{dziugaite2021role} to construct our classifiers. We first use  $Z_{\mathrm{train}}$ to find a prior mean $\bw_{\pi}$. We then set the posterior mean equal to the prior mean $\bw_{\hat{\rho}}=\bw_{\pi}$ but evaluate the r.h.s of the bounds on $Z_{\mathrm{validation}}$. Note that in this way $\Vert \bw_{\hat{\rho}}-\bw_{\pi}\Vert _2^2=0$, while the bound is still valid since the prior is independent of the evaluation set $(X,Y)=Z_{\mathrm{validation}}$. We fit an Isotropic Laplace approximation to the posterior (using $(X,Y)=Z_{\mathrm{validation}}$). This is because for the isotropic Laplace approximation and a Gaussian isotropic prior, the KL divergence has a simple analytical expression
$
\mathrm{KL}(\hat{\rho}\Vert\pi)
=\frac{1}{2}\left(d\frac{\sigma^2_{\hat{\rho}}(\lambda)}{\sigma_{\pi}^2}   + \frac{\Vert \bw_{\hat{\rho}}-\bw_{\pi}\Vert_2 ^2}{\sigma^2_{\pi}} -d-d \ln\sigma^2_{\hat{\rho}}(\lambda)+d \ln\sigma_{\pi}^2 \right)
$
. For different values of $\lambda$ we then estimate the Catoni bound (Theorem \ref{th:catoni}) using $Z_{\mathrm{validation}}$. We also estimate the \emph{test} 0-1 loss. We use the prior variance $\sigma^2_{\pi}=0.1$, as optimizing the marginal likelihood leads to $\sigma^2_{\pi}\approx0$ which is not relevant for BNNs. We plot the results in Figure \ref{catoni_bound_main_fig}. The Catoni bound correlates tightly with test 0-1 loss for all datasets.

\begin{mybox}{blue}{Summary}
The temperature in PAC-Bayes bounds (which is also present in all Chernoff bounds) is a plausible explanation for the need of tempering. It should also be considered first by researchers, before resorting to more involved explanations.
\end{mybox}
\section{Temperature $\lambda$, aleatoric uncertainty and curvature}\label{sec:effect}
\begin{figure*}[t!]
\centering
  \includegraphics[width=\textwidth]{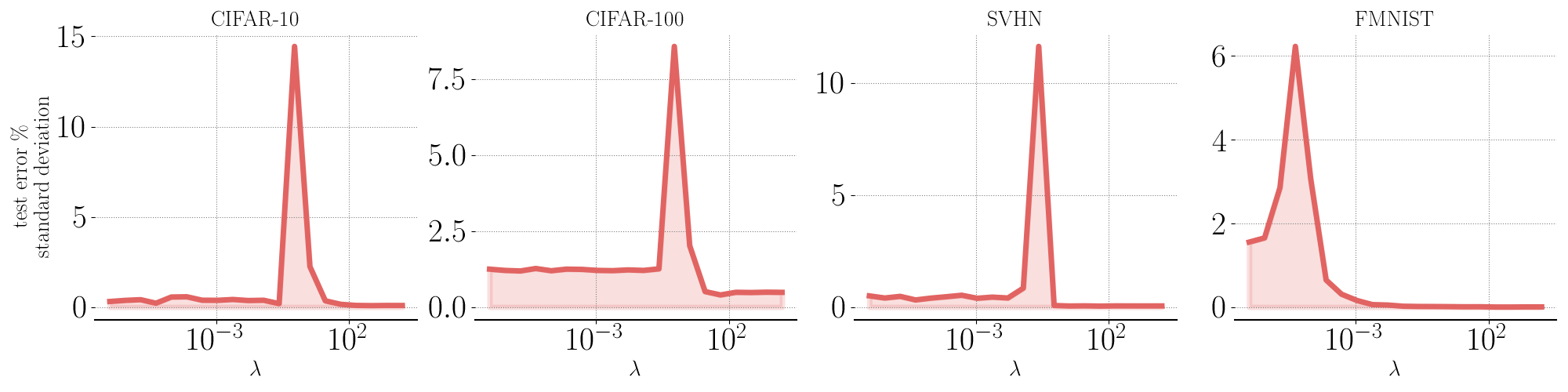}
  \caption{$\lambda$ does not simply fix a misspecified likelihood or prior. \emph{For fixed} $\lambda$ \emph{and noting that the prior} $\sigma^2_{\pi}$ \emph{and the dataset are also fixed} there is large variability in the test 0-1 loss given multiple Laplace approximations. Since the inherent aleatoric uncertainty in this experiment is fixed, we shouldn't be observing any variability in test error for fixed $\lambda$. For $\lambda \gg1$ all the Laplace approximations degenerate to deterministic networks. Conversely, for $\lambda \ll1$ all the Laplace approximations degenerate to random predictions. In both cases, the standard deviation of the error is almost 0.
  }
  \label{map_variability_figure}
\end{figure*}
How does PAC-Bayes align with prior explanations of tempering? In particular, it is interesting to explore the most prominent explanation, which suggests that the temperature $\lambda$ fixes our wrong assumptions about aleatoric uncertainty, encoded in our choice of the likelihood and/or the prior. PAC-Bayes objectives are difficult to analyze theoretically in non-convex cases. Thus in the following we make a number of simplifying assumptions.

First of all we focus our analysis on a linearized model around a minimum of the loss landscape. There are a number of factors supporting this choice. The Laplace approximation with the Generalized Gauss--Newton approximation to the Hessian corresponds to a linearization of the neural network around the MAP estimate $\bw_{\hat{\rho}}\in\mathbb{R}^d$ \citep{immer2021improving} $f_{\mathrm{lin}}(\bx;\bw)=f(\bx;\bw_{\hat{\rho}})+\nabla_{\bw}f(\bx;\bw_{\hat{\rho}})^{\top}(\bw-\bw_{\hat{\rho}})$. Also, prior work has shown that linearization is reasonable when analyzing minima of the loss landscape. This is without having to resort to common assumptions about infinite width \citep{zancato2020predicting,maddox2021fast}.  We adopt the aforementioned linear form together with the Gaussian likelihood $\ell_{\mathrm{nll}}(\bw,\bx,y)=\frac{1}{2}\ln(2\pi\sigma^2)+\frac{1}{2\sigma^2}(y-f_{\mathrm{lin}}(\bx;\bw))^2$. We also make the following modeling choices: 1) Prior over weights: $\bw \sim \mathcal{N}(\bw_{\pi},\sigma_{\pi}^2\bI)$.
2) Gradients as Gaussian mixture: $\nabla_{\bw}f(\bx;\bw_{\hat{\rho}})\sim\sum_{i=1}^k\phi_i\mathcal{N}(\bmu_i,\sigma_{\bx i}^2\bI)$. Note that this assumption should be plausible for \emph{trained} neural networks, in that previous works have shown that per sample gradients with respect to the weights, at $\bw_{\hat{\rho}}$, are clusterable \citep{zancato2020predicting}. We consider that a Gaussian Mixture model for these clusters is a crude but reasonable modeling choice. 
3) Labeling function: $y = f(\bx;\bw_{\hat{\rho}})+\nabla_{\bw}f(\bx;\bw_{\hat{\rho}})^{\top}(\bw_*-\bw_{\hat{\rho}})+\epsilon$, where $\epsilon\sim\mathcal{N}(0,\sigma_{\epsilon}^2)$. 
We also assume that we have a deterministic estimate of the posterior weights $\bw_{\hat{\rho}}$ \emph{which we keep fixed}, and we model the posterior as $\hat{\rho}=\mathcal{N}(\bw_{\hat{\rho}},\sigma^2_{\hat{\rho}}(\lambda)\bI)$, similarly to our experimental section. Therefore estimating the posterior corresponds to estimating the variance $\sigma^2_{\hat{\rho}}(\lambda)$. 
\begin{proposition}[$\mathcal{B}_{\mathrm{approximate}}$]\label{dnn_approximate}
With the above modeling choices, given a distribution $\mathcal{D}$ over $\mathcal{X}\times\mathcal{Y}$, real numbers $\delta \in (0,1]$, $\lambda \in (0,\frac{1}{c})$ with $c = 2 n\sigma^2_{\bx}\sigma^2_{\pi}$, we have  with probability at least $1-\delta$ over  $(X,Y)\sim\mathcal{D}^n$
\begin{equation*}
\begin{split}
&2\bE_{\bw\sim\hat{\rho}}\mathcal{L}_{\mathcal{D}}^{\ell_{\mathrm{nll}}}(\bw) \leq \frac{\Vert \by-f(\bX;\bw_{\hat{\rho}})\Vert ^2_2}{n\mathcolorbox{myred}{\sigma^2}}+\ln(2\pi\mathcolorbox{myred}{\sigma^2})+\sigma_{\epsilon}^2\\
&+\frac{ h }{2n\mathcolorbox{myred}{\sigma^2}}\left(\frac{\mathcolorbox{myblue}{\lambda} h }{d\mathcolorbox{myred}{\sigma^2}}+\frac{1}{\sigma_{\pi}^2}\right)^{-1} 
+\frac{2\sigma_{\bx}^2(\sigma_{\pi}^2d+\Vert \bw_*\Vert _2^2)}{1-2\mathcolorbox{myblue}{\lambda} n \sigma_{\bx}^2\sigma_{\pi}^2}\\
&+\frac{1}{\mathcolorbox{myblue}{\lambda} n} 
\left[
\frac{d}{\sigma_{\pi}^2}  \left(\frac{\mathcolorbox{myblue}{\lambda} h }{d\mathcolorbox{myred}{\sigma^2}} +\frac{1}{2\sigma_{\pi}^2}\right)^{-1}+\frac{\Vert \bw_{\hat{\rho}}-\bw_{\pi}\Vert ^2_2}{\sigma^2_{\pi}} 
-d\right.\\
&\left.+d \ln\left(\frac{\mathcolorbox{myblue}{\lambda} h }{d\mathcolorbox{myred}{\sigma^2}}+\frac{1}{\sigma_{\pi}^2}\right)
+d \ln\sigma_{\pi}^2  
+2\ln\frac{1}{\delta}\right]
\end{split}
\end{equation*}
where $h = \sum_{i}\sum_{j}(\nabla_{\bw}f(\bx_i;\bw_{\hat{\rho}})_j)^2$ is a curvature parameter, $\sigma_{\bx}^2=\sum_{j=1}^k\phi_j\sigma^2_{\bx j}$ is the posterior gradient variance, and $\sigma^2$ is the variance of the likelihood function.
\end{proposition}
Given the Gaussian likelihood function, previous works \citep{bachmann2022tempering,nabarro2021data,aitchison2020statistical} have identified $\sigma^2$ (our estimate of the aleatoric uncertainty of the data) as a key source of misspecification, which the temperature $\lambda$ attempts to fix. Under the PAC-Bayesian modeling of the risk our conclusions are different. Contrary to this prior work our bound suggests that $\lambda$ cannot be seen as simply fixing a misspecified likelihood variance $\sigma^2$ or prior variance $\sigma^2_{\pi}$. In particular it does not simply rescale the aforementioned quantities. Tempering is the result of complex interactions between various parameters resulting from 1) the likelihood through $\sigma^2$, 2) the prior $\pi$ through  $\bw_{\pi}$ and $\sigma_{\pi}^2$, 3) \emph{data} through  $\bw_*$ and $\sigma_{\bx}^2$, 4) as well as various parameters such as the curvature of the minimum $h$ and the MAP estimate $\bw_{\hat{\rho}}$ that depend on the deep neural network architecture, the optimization procedure and the data. Furthermore, our bound implies that \emph{even for fixed prior, likelihood, and data}, the same $\lambda$ can imply different test risks based on each MAP estimates properties such as the curvature $h$ and the distance from initialization $\Vert\bw_{\hat{\rho}}-\bw_{\pi}\Vert_2^2$. 

We can observe this in our empirical data. We first make 10 MAP estimates for the CIFAR-10, CIFAR-100, SVHN and FMNIST datasets and take care such that the training and testing errors of the MAP are similar. We fit for each MAP a Laplace approximation using the Isotropic approximation to the posterior. For a fixed value of $\lambda$ we compute the test 0-1 loss of all the Laplace approximations at the different MAP estimates. We then compute the standard deviation of the test 0-1 loss for this fixed value of $\lambda$. We keep the prior variance fixed $\sigma_{\pi}^2=0.1$. We combine the resulting differences for all fixed values of $\lambda \in [10^{-7},10^{4}]$ in Figure~\ref{map_variability_figure}. 
We see that the standard deviation is large for some fixed values of $\lambda$. \emph{As both the prior, the temperature $\lambda$ and the dataset are kept fixed,} the variability can only be explained in terms of the different posteriors induced by the different MAP estimates. In particular we identify the curvature at the minimum $h$ as important, given that we've taken care such that the training and testing accuracy of the different MAP estimates is similar. Our results imply that the temperature parameter $\lambda$ cannot simply be seen as fixing our wrong initial assumption about the aleatoric uncertainty. 

Assume that different MAP estimates $\bw_{\hat{\rho}}$ are generated from some randomized algorithm $\bw_{\hat{\rho}}\sim\mathscr{P}(X,Y)$ where $X,Y$ are the training data. Our results furthermore imply that it might be difficult to find a single value of $\lambda$ that gives \emph{a target test loss}, simply because of the stochasticity of $\bw_{\hat{\rho}}\sim\mathscr{P}(X,Y)$. This is especially interesting for neural networks because inference is always approximate, and the loss is multimodal. 
\begin{mybox}{blue}{Summary}
$\lambda$ does not simply fix a misspecified estimate of aleatoric uncertainty. In our case, $\lambda$ also varies due to the specific MAP estimate used for Laplace approximation. More generally, $\lambda$ is influenced by the fact that inference is approximate and stochastic.
\end{mybox}
\section{Discussion}
We argued that using Bayesian models to target Frequentist metrics provides a simple explanation for needing a temperature in the optimization objective. We hope that this will motivate Bayesian practitioners to use heuristics ``guilt-free" when targeting Frequentist performance metrics, or target posterior contraction instead. We showcased the importance of choosing a metric when discussing tempering, and our empirical results on the tradeoff between metrics point towards the need for the use of Pareto curves when evaluating Bayesian approaches. Finally our results on the effect of the posterior on tempering might imply that some fine-tuning of $\lambda$ might be inevitable given the approximate and stochastic nature of inference in neural networks.

\clearpage

\bibliography{acml23}

\end{document}